\documentclass[letterpaper]{article} 
\usepackage{aaai25}  
\usepackage{times}  
\usepackage{helvet}  
\usepackage{courier}  
\usepackage[hyphens]{url}  
\usepackage{graphicx} 
\usepackage{natbib}  
\usepackage{caption} 
\usepackage{algorithm}
\usepackage{algorithmic}

\usepackage{newfloat}
\usepackage{listings}
\DeclareCaptionStyle{ruled}{labelfont=normalfont,labelsep=colon,strut=off} 
\floatstyle{ruled}
\newfloat{listing}{tb}{lst}{}
\floatname{listing}{Listing}
\usepackage{subcaption}
\usepackage{multirow}
\usepackage{tabularray}
\usepackage{booktabs} 
\usepackage{amssymb}
\usepackage{xcolor}
\usepackage{amsmath}

\title{Spatiotemporal Blind-Spot Network with Calibrated Flow Alignment for Self-Supervised Video Denoising}
\author{
    Zikang Chen,
    Tao Jiang,
    Xiaowan Hu,
    Wang Zhang,
    Huaqiu Li,
    Haoqian Wang\thanks{Corresponding author.}
}
\affiliations{
    Shenzhen International Graduate School, Tsinghua University

    \{czk23,jiang-t23,hu-xw19,zhangwan23,lihq23\}@mails.tsinghua.edu.cn,
    wanghaoqian@tsinghua.edu.cn
}

\usepackage{bibentry}

\begin{document}

\maketitle

\begin{abstract}
Self-supervised video denoising aims to remove noise from videos without relying on ground truth data, leveraging the video itself to recover clean frames. Existing methods often rely on simplistic feature stacking or apply optical flow without thorough analysis. This results in suboptimal utilization of both inter-frame and intra-frame information, and it also neglects the potential of optical flow alignment under self-supervised conditions, leading to biased and insufficient denoising outcomes. To this end, we first explore the practicality of optical flow in the self-supervised setting and introduce a \textbf{S}patio\textbf{T}emporal \textbf{B}lind-spot \textbf{N}etwork (\textbf{STBN}) for global frame feature utilization. In the temporal domain, we utilize bidirectional blind-spot feature propagation through the proposed blind-spot alignment block to ensure accurate temporal alignment and effectively capture long-range dependencies. In the spatial domain, we introduce the spatial receptive field expansion module, which enhances the receptive field and improves global perception capabilities. Additionally, to reduce the sensitivity of optical flow estimation to noise, we propose an unsupervised optical flow distillation mechanism that refines fine-grained inter-frame interactions during optical flow alignment. Our method demonstrates superior performance across both synthetic and real-world video denoising datasets. The source code is publicly available at https://github.com/ZKCCZ/STBN.
\end{abstract}

%

\begin{figure}[t]
\centering
\includegraphics[width=0.98\columnwidth]{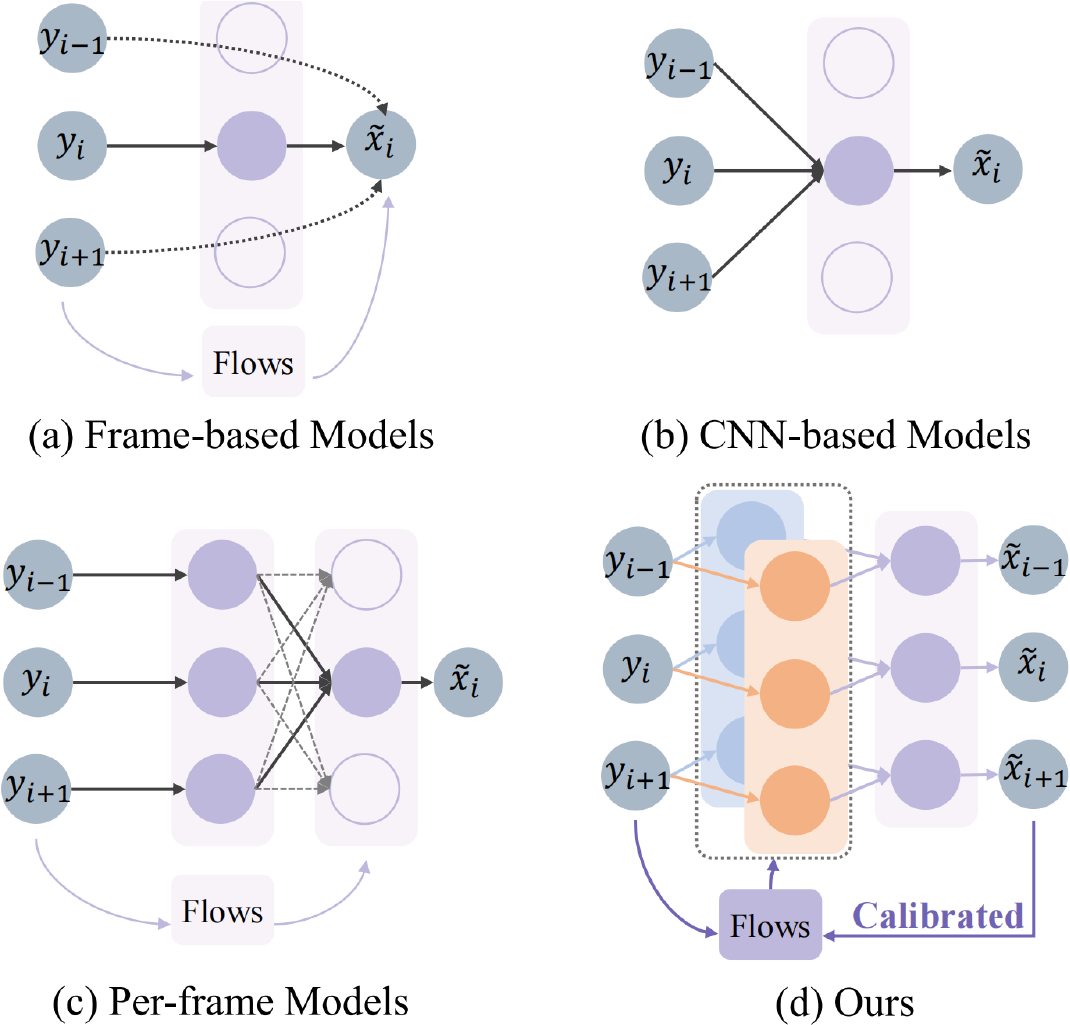} 
\caption{Illustrative comparison of frame sequence utilization strategies in self-supervised video denoising methods.}
\label{fig1}
\end{figure}

\section{Introduction}
Images captured under challenging environmental conditions, such as low lighting and slow shutter speeds, are often susceptible to various forms of noise and corruption. This issue is exacerbated in videos due to the typically higher shutter speeds, which not only degrades the overall quality of the video but also adversely affects subsequent computer vision tasks~\cite{shen2020noise,deng2022nightlab}.

Given its critical role in computer vision, video denoising has witnessed significant advancements, largely driven by the application of deep learning techniques. Supervised video denoising methods, including Convolutional Neural Networks (CNNs)~\cite{tassano2019dvdnet,tassano2020fastdvdnet}, Recurrent Neural Networks (RNNs)~\cite{chan2021basicvsr,li2022unidirectional}, and Transformer-based models~\cite{liang2022recurrent,liang2024vrt}, have made significant advancements. However, supervised video denoising methods rely heavily on labeled data, which is difficult and time-consuming to obtain. For example, obtaining the ground truth data of microscope videos and dynamic scenes is often impractical. This limitation restricts the applicability of supervised approaches in these contexts.
Therefore, self-supervised methods have gained increasing attention as they eliminate the need for labeled training data. Grounded in the Noise2Noise assumption~\cite{lehtinen2018noise2noise}, frame-based approaches~\cite{ehret2019model,dewil2021self} warp consecutive frames to create noise pairs for self-supervised training, as illustrated in Figure~\ref{fig1}a. These methods heavily rely on precise optical flow estimation, which becomes particularly challenging in high-noise scenarios. The dependency can lead to severe artifacts in the warped images and an inefficient utilization of inter-frame redundancy. Additionally, CNN-based models~\cite{sheth2021unsupervised}, depicted in Figure~\ref{fig1}b, stack adjacent frames and employ blind-spot networks for self-supervised training. Per-frame models, as shown in Figure~\ref{fig1}c, attempt to leverage all other aligned frames for each frame. However, this results in a computational complexity of $O(T^2)$. One possible approach is to align only a few adjacent frames~\cite{zheng2023unsupervised}, yet this still compromises long-term information. These models are limited by their frame window size, restricting their ability to capture global temporal information.

Apart from the limited receptive field in both spatial and temporal domains, another significant issue lies in the efficiency and accuracy of optical flow utilization. The aforementioned methods that rely on frame-by-frame optical flow matching encounter a high computational complexity. Methods like RDRF~\cite{wang2023recurrent} tackle these challenges by recurrently leveraging optical flow to capture long-term dependencies. However, as noted in their approach, their model is prone to overfitting, especially when dealing with real-world noisy data. Moreover, the reliance on unverified optical flow can introduce potential biases and errors, which need to be carefully examined within a self-supervised framework. Additionally, current methods are restricted to only access corrupted input video sequences for optical flow estimation, leading to suboptimal results due to the noise sensitivity of optical flow estimation.

To address the aforementioned challenges, we introduce a Spatiotemporal Blind-spot Network (STBN) to robustly handle both synthetic and real-world noise, as shown in Figure~\ref{fig1}d. Our approach leverages inter-frame information through bidirectional alignment and propagation with the Blind-Spot Alignment (BSA) block for global temporal awareness. To integrate aligned temporal information and intra-frame features, we propose the Spatial Receptive Field Expansion (SRFE) module, which significantly enlarges the receptive field and further utilizes bidirectional spatial information. In the self-supervised setting, we discuss and calibrate feature alignment methods to ensure the consistency of noise distribution and independence, preserving the integrity of our self-supervised assumptions and avoiding potential biases. Moreover, considering the sensitivity of optical flow estimation to noise, we perform optical flow refinement using initially restored frames as pseudo-ground truth for knowledge distillation, enhancing noise robustness and improving spatiotemporal feature alignment and utilization. We summarize our contributions as follows:

\begin{itemize}
\item We propose a Spatiotemporal Blind-spot Network that effectively leverages inter-frame and intra-frame information through blind-spot temporal propagation and spatial fusion for self-supervised denoising for both synthetic and real noise.

\item To ensure accurate utilization of temporal information in our self-supervised framework, we calibrate the multi-frame alignment paradigm to maintain global consistency of noise priors to prevent bias during training.

\item The proposed knowledge distillation strategy in an unsupervised setting mitigates the sensitivity of optical flow to noise, thereby enhancing the precision of spatiotemporal feature utilization.

\item Experimental results show that our method surpasses existing state-of-the-art self-supervised methods on various synthetic and real video noise datasets, demonstrating its superiority in video denoising tasks.
\end{itemize}

\begin{figure*}[t]
\centering
\includegraphics[width=1\textwidth]{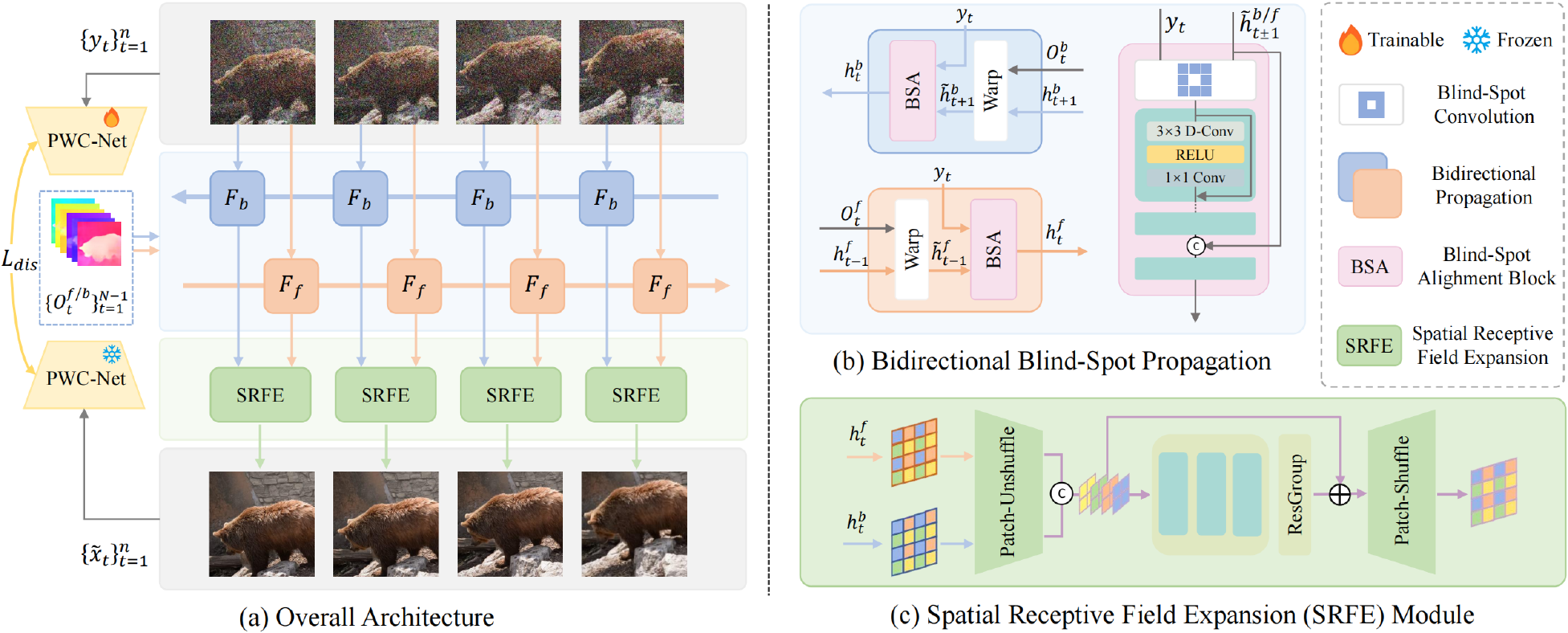} %
\caption{Illustration of the proposed method: (a) Overall architecture of STBN, including spatiotemporal feature aggregation and optical flow refinement. (b) The Bidirectional Blind-Spot Propagation utilizes the BSA block for global temporal awareness in both forward and backward propagation. (c) Detailed process of the Spatial Receptive Field Expansion module, which sequentially incorporates patch-shuffle, residual blocks, and patch-unshuffle to effectively enhance the spatial receptive field.}
\label{arch}
\end{figure*}

\section{Related Work}
\subsection{Supervised Video Denoising}
To leverage temporal redundancy to exploit inter-frame information, methods such as PaCNet~\cite{ko2018pac} and VNLNet~\cite{davy2019non} utilize block matching combined with CNNs based on spatiotemporal neighborhoods, leading to high computational complexity. Alternatively, sliding window approaches like FastDVDnet~\cite{tassano2020fastdvdnet}, an extension of DVDnet~\cite{tassano2019dvdnet}, enhance efficiency by processing fixed-size consecutive frames through a two-level U-Net. Some methods incorporate optical flow for motion compensation, such as FloRNN~\cite{li2022unidirectional}, which extends BasicVSR~\cite{chan2021basicvsr} by integrating future frame alignment for online denoising. VRT~\cite{liang2024vrt} processes video sequences in 2-frame clips with attention modules and optical flow for cross-clip interactions. RVRT~\cite{liang2022recurrent} further enhances this by processing frames in parallel within a global recurrent framework. 

\subsection{Unsupervised Video Denoising}

Traditional methods, such as VBM4D~\cite{maggioni2012video} based on BM3D~\cite{dabov2007color}, use video filtering algorithms to find similar blocks for denoising. Recent deep learning-based approaches can be broadly categorized into noise-paired methods and blind-spot network methods. Frame2Frame (F2F)~\cite{ehret2019model} and Multi-Frame2Frame (MF2F)~\cite{dewil2021self}, based on the Noise2Noise (N2N)~\cite{lehtinen2018noise2noise} assumption, align consecutive frames as noise pairs for denoising.
ER2R~\cite{zheng2023unsupervised} extends the R2R~\cite{pang2021recorrupted} assumption, training by creating noise pairs through adding and subtracting noise from the original noisy videos when the specific noise distribution is known. It aligns each frame with others using a sliding window to reduce complexity, which leads to a significant loss of temporal information.
Another approach extends blind-spot networks~\cite{krull2019noise2void,laine2019high} to the video denoising domain. UDVD~\cite{sheth2021unsupervised} directly stacks a fixed length of adjacent frames into a blind-spot CNN. Although this method implicitly achieves feature alignment through a two-stage U-Net, it restricts the ability to utilize long-term temporal patterns by considering only frames within a limited window size. RDRF~\cite{wang2023recurrent} employs 3D networks and a recurrent network based on blind spatial modulation to integrate features from near and far. However, this method is prone to overfitting, especially when dealing with raw video data.

\subsection{Frame Alignment in Video Restoration}
In video restoration, aligning highly correlated but temporally unsynchronized frames is crucial~\cite{nah2019recurrent,chan2021basicvsr}. Many methods use optical flow for frame alignment. BasicVSR~\cite{chan2021basicvsr} employs optical flow for recurrent feature propagation, and BasicVSR++~\cite{chan2022basicvsr++} uses it to guide offset learning. Task-specific optical flow is fine-tuned using models like SpyNet~\cite{ranjan2017optical} and PWC-Net~\cite{sun2018pwc} for specific restoration tasks~\cite{xue2019video}. Despite its efficiency in video restoration~\cite{chan2021basicvsr,chan2022basicvsr++}, the use of optical flow in self-supervised denoising has been less explored. UDVD~\cite{sheth2021unsupervised} achieves implicit alignment with a two-stage U-Net, while some methods~\cite{yu2020joint} use trainable estimators for improved alignment. The applicability and effectiveness of optical flow alignment in self-supervised denoising remain to be explored.

\section{Methodology}
Let $\boldsymbol{y}\in\mathbb{R}^{T\times H\times W\times C}$ represent the noisy input frame sequence and $\boldsymbol{x}\in\mathbb{R}^{T\times H\times W\times C}$  denote the potentially clean target frame sequence, where $T$, $H$, $W$, and $C$ are the video length, height, width, and channel, respectively. The overall framework of STBN is illustrated in Figure~\ref{arch}a. Initially, optical flow is predicted from the noisy video sequence and fed into the bidirectional blind-spot propagation module, where features are aligned within the Blind-Spot Alignment (BSA) block. The temporal information is then passed to the Spatial Receptive Field Expansion (SRFE) module, significantly expanding the receptive field of the blind-spots. The fused features are used to generate the final output and serve as pseudo-ground truth for further optical flow refinement.

\subsection{Calibration of Frame Alignment}
To achieve global temporal feature utilization, we employ optical flow for bidirectional feature warping. In this section, we examine the applicability of optical flow alignment methods within the self-supervised learning framework. 

\begin{figure}[t!]
    \centering
    \begin{subfigure}[b]{0.73\columnwidth}  
        \centering
        \includegraphics[width=\textwidth]{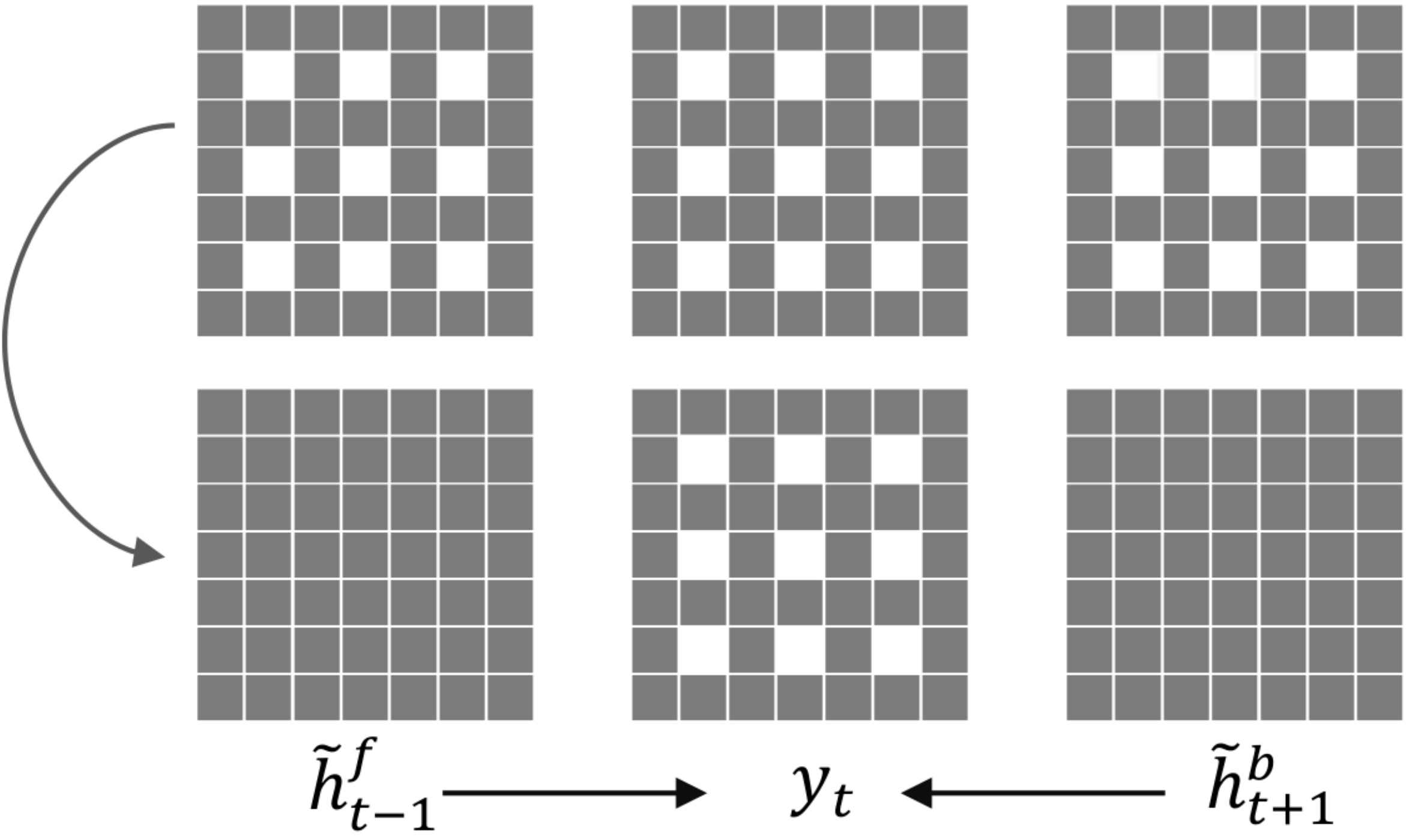}
        \caption{}
        \label{recf1}
    \end{subfigure}
    \hfill  
    \begin{subfigure}[b]{0.18\columnwidth}  
        \centering
        \includegraphics[width=\textwidth]{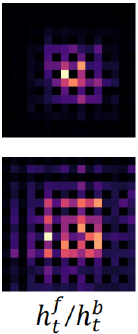}
        \caption{}
        \label{recf2}
    \end{subfigure}

    \caption{Visualization of (a) BSA block for temporal processing and (b) SRFE for spatial receptive field expansion.}
    \label{recf}
\end{figure}

First, we propose the blind-spot network assumption for video sequences, where noise is pixel-independent both temporally and spatially, and pixel information can be inferred from the spatiotemporal context in the video. We assume that at the $t$-th frame $\boldsymbol{y}_t$, the receptive field for the $i$-th pixel $\boldsymbol{y}_{(t,i)}$, which acts as the blind-spot in our model, is denoted as $\boldsymbol{y}_{t,RF(i)}$. We define our model as the function as follows:
\begin{multline}
f\big(\boldsymbol{y}_{t,RF(i)},warp(\boldsymbol{y}_k,\boldsymbol{O}_{k});\boldsymbol{\theta}\big) = \boldsymbol{y}_{(t,i)}, \\
k \in \{1,2,...,T\}\setminus\{t\},
\end{multline}
where 
$\boldsymbol{\theta}$ denotes the vector of model parameters we aim to train, $\boldsymbol{O}_{k}$ represents the estimated optical flow between two frames, and 
$warp$ represents the alignment operation. The model is trained by minimizing the empirical risk below:
\begin{equation}
\mathop{\arg\min}_\theta\sum_{t,i}L\left(f\left(\boldsymbol{y}_{(t,RF(i))},warp(\boldsymbol{y}_k,\boldsymbol{O}_{k});\boldsymbol{\theta}\right),\boldsymbol{y}_{(t,i)}\right).
\end{equation}
The above formulation can be considered equivalent to the supervised training process. The detailed proof is provided in the supplementary material. 

\begin{figure}[t]
\centering
\includegraphics[width=1\columnwidth]{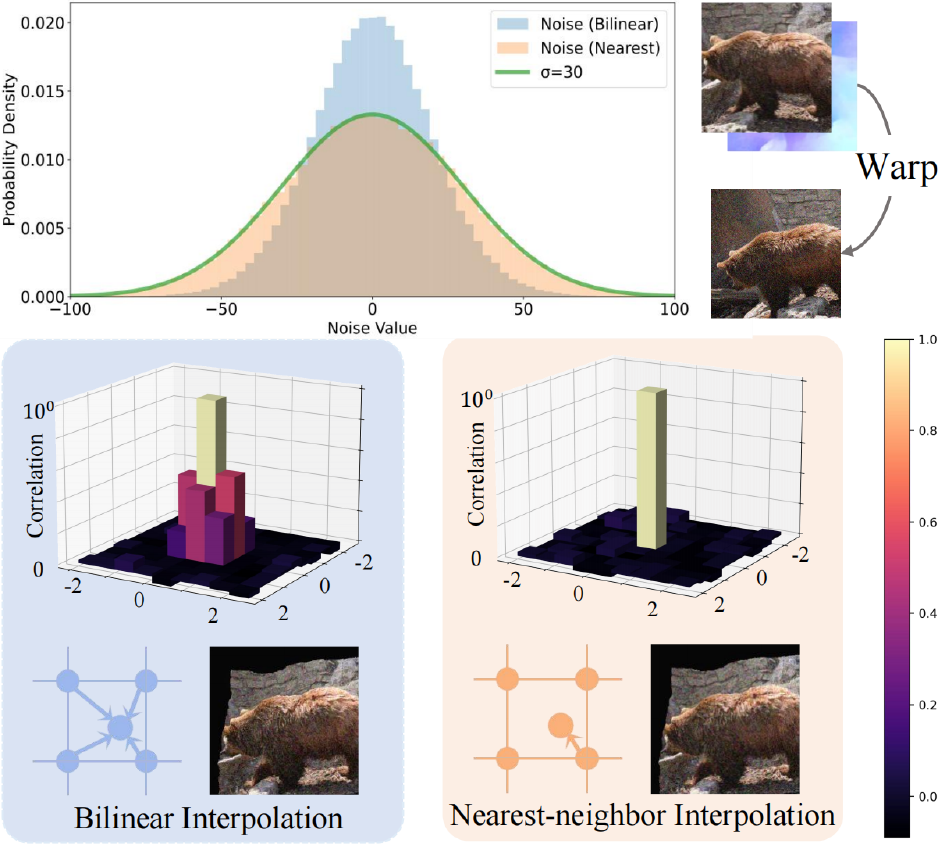}
\caption{Visualization of noise distribution and correlation for two interpolation methods. Bilinear interpolation introduces spatial correlation and distorts the noise distribution, while nearest-neighbor interpolation preserves it.}
\label{flow_warp}
\end{figure}

As shown in the above derivation, the inputs to $f$ necessitate that both $\boldsymbol{y}_t$ and $warp(\boldsymbol{y}_k,\boldsymbol{O}_{k})$, i.e., the noise from the current frame and the aligned frames, must remain pixel-independent both temporally and spatially.
In optical flow alignment, bilinear and nearest-neighbor interpolation are two commonly employed methods. We use these as examples to illustrate the impact of alignment on noise characteristics and correlation. As shown in Figure~\ref{flow_warp}, we performed forward warping on frames using these two methods, respectively. The same operation is applied on the ground truth data to calculate the noise distribution after interpolation. It can be observed that bilinear interpolation not only disrupts the distribution of noise but also introduces spatial correlations. This occurs because bilinear interpolation uses surrounding pixel information, performing a filtering-like operation on the image, which violates our self-supervised assumptions and leads to method failure. In contrast, nearest-neighbor interpolation preserves the original pixel values, maintaining the noise distribution and its independence. This is further demonstrated in our experiments.

\begin{figure*}[t]
\centering
\includegraphics[width=1\textwidth]{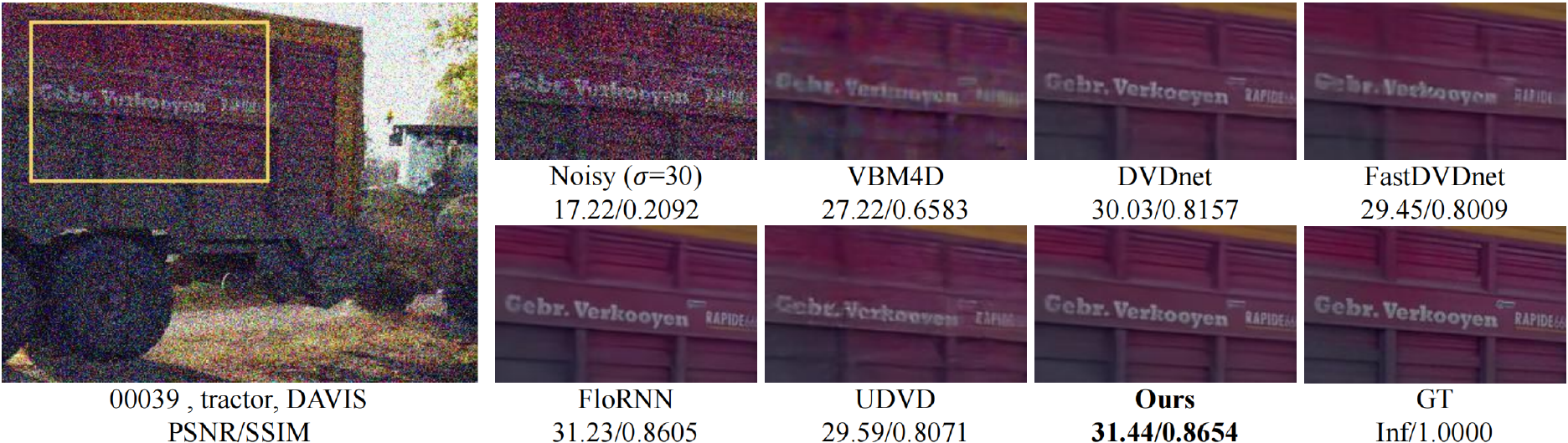} 
\caption{Visual comparisons of different methods on synthetic noise data.}
\label{vis:davis}
\end{figure*}

\begin{table*}[!ht]
  \centering
  \resizebox{1\linewidth}{!} {
  
  \renewcommand{\arraystretch}{1.1} 
  
  \begin{tabular}{c|c|c|ccc|cccc}
    \toprule
    \multirow{2}{*}{Dataset} & \multirow{2}{*}{$\sigma$} & \multicolumn{1}{c|}{Traditional} & \multicolumn{3}{c|}{Supervised} & \multicolumn{4}{c}{Unsupervised} \\
    & & \multicolumn{1}{c|}{VBM4D} & \multicolumn{1}{c}{DVDnet} & \multicolumn{1}{c}{FastDVDnet} & \multicolumn{1}{c|}{FloRNN} & \multicolumn{1}{c}{UDVD} & \multicolumn{1}{c}{RDRF} & \multicolumn{1}{c}{ER2R$_s$} & \multicolumn{1}{c}{\textbf{STBN (Ours)}} \\
    \hline
    \multirow{6}{*}{\textbf{Set8}}  
    & 10       & 36.05/- & 36.08/0.9510 & 36.44/0.9540 & 37.57/0.9639 & 36.36/0.9510 & 36.67/0.9547 & 37.55/- & \textbf{37.24/0.9594} \\
    & 20       & 32.19/- & 33.49/0.9182 & 33.43/0.9196 & 34.67/0.9379 & 33.53/0.9167 & 34.00/0.9251 & 34.34/- & \textbf{34.41/0.9322} \\
    & 30       & 30.00/- & 31.68/0.8862 & 31.68/0.8889 & 32.97/0.9138 & 31.88/0.8865 & 32.39/0.8978 & 32.45/- & \textbf{32.76/0.9072} \\
    & 40       & 28.48/- & 30.46/0.8564 & 30.46/0.8608 & 31.75/0.8911 & 30.72/0.8595 & 31.23/0.8725 & 31.09/- & \textbf{31.57/0.8837} \\
    & 50       & 27.33/- & 29.53/0.8289 & 29.53/0.8351 & 30.80/0.8696 & 29.81/0.8349 & 30.31/0.8490 & 30.05/- & \textbf{30.62/0.8608} \\
    \cline{2-10}
    & avg      & 30.81/- & 32.29/0.8881 & 32.31/0.8917 & 33.55/0.9153 & 32.46/0.8897 & 32.92/0.8998 & 33.10/- & \textbf{33.32/0.9087} \\
    \hline
    \multirow{6}{*}{\textbf{DAVIS}} 
    & 10       & 37.58/- & 38.13/0.9657 & 38.71/0.9672 & 40.16/0.9755 & 39.17/0.9700 & 39.54/0.9717 & 39.52/- & \textbf{40.35/0.9613} \\
    & 20       & 33.88/- & 35.70/0.9422 & 35.77/0.9405 & 37.52/0.9564 & 35.94/0.9428 & 36.40/0.9473 & 36.49/- & \textbf{37.67/0.9606} \\
    & 30       & 31.65/- & 34.08/0.9188 & 34.04/0.9167 & 35.89/0.9440 & 34.09/0.9178 & 34.55/0.9245 & 34.60/- & \textbf{36.00/0.9454} \\
    & 40       & 30.05/- & 32.86/0.8962 & 32.82/0.8949 & 34.66/0.9286 & 32.79/0.8949 & 33.23/0.9032 & 33.29/- & \textbf{34.73/0.9296} \\
    & 50       & 28.80/- & 31.85/0.8745 & 31.86/0.8747 & 33.67/0.9131 & 31.80/0.8739 & 32.20/0.8832 & 32.25/- & \textbf{33.70/0.9138} \\
    \cline{2-10}  
    & avg      & 32.39/- & 34.52/0.9195 & 34.64/0.9188 & 36.38/0.9435 & 34.76/0.9199 & 35.18/0.9260 & 35.23/- & \textbf{36.49/0.9451} \\
    \bottomrule
  \end{tabular}}
  
  \caption{Quantitative comparison of PSNR/SSIM for Gaussian denoising on the Set8 and DAVIS datasets. The best results for unsupervised methods are in bold. Note that ER2R$_s$ utilizes the same video sequence for both training and testing.}
  \label{tab:gauss}
\end{table*}

\subsection{Spatiotemporal Blind-Spot Feature Aggregation}
To better utilize video frame sequences in both spatial and temporal domains, we design two distinct modules: the temporal module, which performs bidirectional alignment and propagation of features, and the spatial module, which significantly expands the receptive field to more effectively leverage the aligned frames. Together, these modules enable the model to achieve global awareness and enhance spatiotemporal feature integration.
\subsubsection{Bidirectional Blind-Spot Propagation.}
To perform temporal feature alignment and propagation, we design a feature propagation and alignment module using blind-spot convolutions and dilated convolutions, as illustrated in Figure~\ref{arch}b. The input $\boldsymbol{y}_t$ from the $t$-th frame, along with the bidirectionally propagated features $\boldsymbol{h}_{t-1}^{f}$ or $\boldsymbol{h}_{t+1}^{b}$, which are warped to the current frame using optical flow, are then fed into the Blind-Spot Alignment (BSA) block for motion compensation. The entire process is as follows:
\begin{equation}
\begin{aligned}
\boldsymbol{h}_t^f=F_f\left(\boldsymbol{y}_{t},warp(\boldsymbol{h}_{t-1}^f,\boldsymbol{O}_{t}^f)\right),\\
\boldsymbol{h}_t^b=F_b\left(\boldsymbol{y}_{t},warp(\boldsymbol{h}_{t+1}^b,\boldsymbol{O}_{t}^b)\right),
\end{aligned}
\end{equation}
where $F_f$, $F_b$ denote forward and backward propagation, $\boldsymbol{O}_{t}^f$, $\boldsymbol{O}_{t}^b$ represent the bidirectional estimated optical flow. 

During the alignment process, the BSA block is designed to maximally leverage the features from both forward and backward propagation. First, \(\boldsymbol{y}_t\) and \(\boldsymbol{h}\) are concatenated and then passed through a blind-spot convolution. The output is subsequently processed by modules that consist of a dilated convolution, an activation layer, and a \(1 \times 1\) convolution. Although the features are well-aligned at this stage, they are not fully utilized. Therefore, we further concatenate the output with feature $\boldsymbol{h}$ and pass it through the blind-spot convolution block. Figure~\ref{recf1} shows the dependency between input and output pixels, with white pixels indicating regions independent of the central pixel and gray pixels representing convolution weights. This demonstrates that the BSA block effectively utilizes all temporal redundancy.

\subsubsection{Spatial Receptive Field Expansion.}
Once the bidirectional features are aligned to \(\boldsymbol{y}_t\), they inherently capture the temporal features of the entire sequence. To further leverage the aligned frames, we expand the receptive field under the blind-spot framework to utilize spatial domain information for enhanced image recovery. Inspired by ~\cite{jang2024puca,li2024tbsn}, we propose the Spatial Receptive Field Expansion (SRFE) Module, as illustrated in Figure~\ref{arch}c.

In the SRFE module, the forward features \(\boldsymbol{h}^f\) and backward features \(\boldsymbol{h}^b\) are first processed through a patch-unshuffle operation, and then stacked together to pass through several residual blocks, which ensure thorough feature fusion and enhance the model's ability to capture contextual information. Finally, the features are restored to their original size through a patch-shuffle operation, producing the output. The process can be represented as follows:
\begin{equation}
\boldsymbol{\tilde{x}}_t=\mathbf{SRFE}(\boldsymbol{h}_t^b,\boldsymbol{h}_t^f),~t=1,2,...T
\end{equation}
As shown in Figure~\ref{recf2}, our strategy leads to a substantial increase in the receptive field, effectively integrating spatial information. This expansion enhances the model's capability to capture and utilize detailed spatial features.

\subsection{Flow Refinement in Noise}
Optical flow estimation, which is also sensitive to noise, significantly impacts the accuracy of temporal alignment. To address the issue of imprecise optical flow estimation caused by corrupted input images, we introduce a knowledge distillation approach that uses pseudo-ground truth for optical flow refinement. 

We define the method of optical flow estimator as $\mathcal{E}(\cdot)$. Once the training achieves preliminary effectiveness, we generate clean video sequences $\tilde{x}$ using a frozen-parameter \(\mathcal{E}_{\text{fix}}(\cdot)\) to serve as pseudo-ground truth. These sequences and the original video frame $\boldsymbol{y}$ are used for optical flow estimation respectively as follows:
\begin{equation}
\boldsymbol{\tilde{O}}_{t}^f=sg(\mathcal{E}_{\text{fix}}(\boldsymbol{\tilde{x}}_{t},\boldsymbol{\tilde{x}}_{t+1})),
~\boldsymbol{O}_{t}^f=\mathcal{E}(\boldsymbol{y}_{t},\boldsymbol{y}_{t+1}),
\end{equation}
where $sg(\cdot)$ is the stop gradient operation. This accurate optical flow $\boldsymbol{\tilde{O}}_{t}^f$ treated as pseudo-ground truth is then to guide the refinement of the optical flow estimation in noisy video sequences. We optimize the original optical flow estimator using the following loss function:
\begin{equation}
\mathcal{L}_{dis} = \sum_{t} \left\|\boldsymbol{\tilde{O}}_{t}^f - \boldsymbol{O}_{t}^f\right\|_1.
\end{equation}
The distillation loss, scaled by a small coefficient $\alpha$ as a constraint, is jointly trained with our model. This distillation approach enhances the performance of the optical flow estimator in the presence of noise, thereby improving overall temporal alignment and benefiting the entire model.

\begin{figure*}[t]
\centering
\includegraphics[width=1\textwidth]{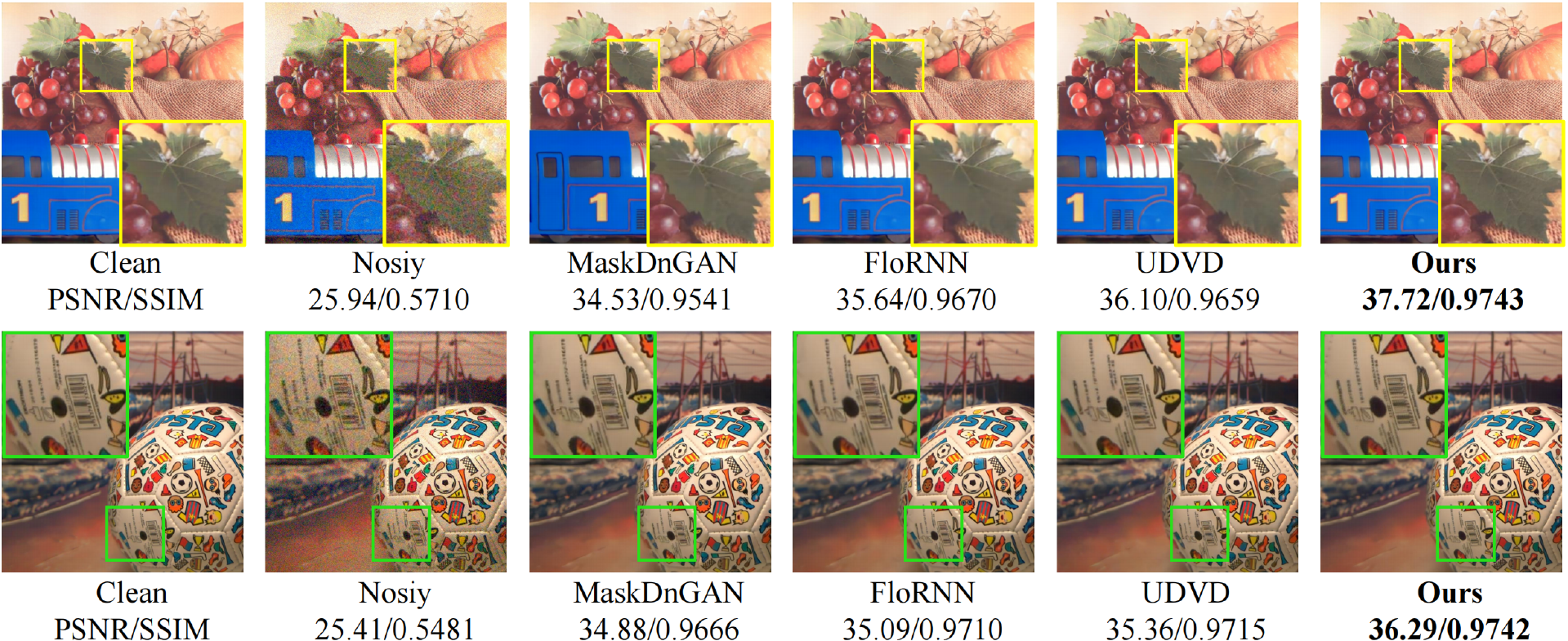} 
\caption{Visual comparisons on CRVD dataset. The results have been converted to the sRGB domain for visualization.}
\label{vis:raw}
\end{figure*}

\begin{table*}[ht]
  \centering
  \setlength{\tabcolsep}{1.4mm} %
  \renewcommand{\arraystretch}{1.2} 
  \begin{tabular}{c|cccc|cccc}
    \toprule
    \multirow{2}{*}{ISO} & \multicolumn{4}{c|}{Supervised} & \multicolumn{4}{c}{Unsupervised} \\
    & \multicolumn{1}{c}{FastDVDnet} & RViDeNet & MaskDnGAN & \multicolumn{1}{c|}{FloRNN} & UDVD & RDRF & ER2R$_p$ & \textbf{STBN (Ours)} \\
    \hline
          1600  & 43.43/0.9866 & 47.74/0.9938 & 47.52/0.9941 & 48.81/0.9956 & 48.02/0.9982 & 48.38/0.9983 & 49.14/- & \textbf{49.27/0.9988} \\
          3200  & 42.91/0.9844 & 45.91/0.9911 & 45.88/0.9914 & 47.05/0.9933 & 46.44/0.9980 & 46.86/0.9981 & 47.51/- & \textbf{47.58/0.9985} \\
          6400  & 40.29/0.9793 & 43.85/0.9880 & 44.14/0.9886 & 45.09/0.9910 & 44.74/0.9972 & 45.24/0.9975 & 45.61/- & \textbf{45.75/0.9980} \\
          12800 & 36.05/0.9613 & 41.20/0.9819 & 41.48/0.9834 & 42.63/0.9866 & 42.21/0.9966 & 42.72/0.9969 & 43.03/- & \textbf{43.36/0.9976} \\
          25600 & 36.50/0.9400 & 41.17/0.9821 & 40.79/0.9819 & 42.19/0.9872 & 42.13/0.9951 & 42.25/0.9948 & 42.91/- & \textbf{42.91/0.9972} \\
    \midrule
    avg   & 39.84/0.9703 & 43.97/0.9874 & 43.96/0.9880 & 45.15/0.9907 & 44.71/0.9970 & 45.09/0.9971 & 45.64/- & \textbf{45.77/0.9980} \\
    \bottomrule
  \end{tabular}
\caption{Quantitative comparison of PSNR/SSIM on the CRVD dataset. The best results for unsupervised methods are in bold. Note that ER2R$_p$ utilizes extra noise distribution priors to generate noise pairs during the training process.}
\label{tab:raw}
\end{table*}

\section{Experiments}
\subsection{Implementation Details}
We conduct experiments on both synthetic and real raw noise. For synthetic noise, following~\cite{sheth2021unsupervised,wang2023recurrent}, we train our model with negative log-likelihood loss $\mathcal{L}_{log}$ and test them with posterior inference~\cite{laine2019high}. For real raw noise, we use $\mathcal{L}_2$ loss for self-supervised training. 
For the optical flow estimator, we use the pre-trained PWC-Net~\cite{sun2018pwc} as our initial optical flow extractor. The distillation loss is introduced with $\alpha=5 \times 10^{-4}$. Training sequences are spatially cropped to a size of \(96 \times 96\) and temporally to a length of \(T = 10\) for synthetic data and \(T = 7\) for real data. All experiments are carried out using the Adam optimizer with an initial learning rate of $1 \times 10^{-4}$ on a single RTX 3090 GPU. We used Peak Signal-to-Noise Ratio (PSNR) and Structural Similarity (SSIM) as evaluation metrics.

\subsection{Experiments on Synthetic Noise}
In our experiments on synthetic noise, we utilize DAVIS dataset~\cite{pont20172017} and Set8~\cite{tassano2019dvdnet} dataset.  
To generate noisy video sequences, additive white Gaussian noise (AWGN) with a standard deviation ${\sigma \in [5,55]}$ is introduced to the training dataset.
We compare our method with a range of benchmarks, including the non-learning method VBM4D~\cite{maggioni2012video}, supervised approaches such as FastDVDnet~\cite{tassano2020fastdvdnet}, PaCNet~\cite{vaksman2021patch}, and FloRNN~\cite{li2022unidirectional}, as well as unsupervised methods like UDVD~\cite{sheth2021unsupervised}, RDRF~\cite{wang2023recurrent}, and ER2R~\cite{zheng2023unsupervised}. 

\subsubsection{Quantitative Comparison.} 
Table \ref{tab:gauss} reports the PSNR and SSIM of different methods on the DAVIS testing set and Set8 datasets under different noise levels. Note that ER2R utilizes the same video sequence for both training and testing. Our model outperforms RDRF by an average PSNR of 1.31 dB and 0.4 dB on two different datasets and is highly comparable to the supervised method FloRNN. The results demonstrate the effectiveness of our global spatiotemporal perception and refined optical flow alignment, highlighting the advantages of our self-supervised method. Figure~\ref{vis:davis} illustrates our qualitative results, showing that our method restores corrupted text more accurately compared to existing approaches, which demonstrates the effectiveness of our approach in preserving fine details.

\subsection{Experiments on Real Raw Noise}
We evaluate our method using the CRVD dataset~\cite{yue2020supervised}, a real-world video denoising dataset captured in the raw domain, to assess our performance on real-world noise. This dataset comprises 6 indoor scenes for training and 5 indoor scenes for testing, with each scene consisting of 7 frames with 10 different noise realizations captured at five different ISO levels. 
We compare our method against several approaches, including supervised methods FastDVDnet~\cite{tassano2020fastdvdnet}, RViDeNet~\cite{yue2020supervised}, MaskDnGAN~\cite{paliwal2021multi}, and FloRNN~\cite{li2022unidirectional}, as well as unsupervised methods such as UDVD~\cite{sheth2021unsupervised}, RDRF~\cite{wang2023recurrent}, and ER2R~\cite{zheng2023unsupervised}. For a fair comparison, both training and testing are performed on the test sequences as employed in previous unsupervised methods.
\subsubsection{Quantitative Comparison.} 
Table \ref{tab:raw} reports our results on the CRVD dataset. Note that ER2R utilizes prior noise information by creating noise pairs during training, whereas ours rely solely on noisy images. Our model demonstrates superior performance, surpassing RDRF by 0.68 dB under identical settings. While RDRF requires meticulous tuning to prevent overfitting with limited samples, our method leverages well-calibrated optical flow alignment and a robust spatiotemporal blind-spot network, which enables precise global information aggregation to improve denoising results. Our approach exceeds ER2R by an average of 0.13 dB even though we have very limited data. Under the above training setting, we outperform the supervised method FloRNN by an average of 0.62 dB. Given the challenges of obtaining ground truth in real noise scenarios, our unsupervised approach demonstrates greater practical utility. As illustrated in Figure \ref{vis:raw}, our method better preserves high-frequency details that others often lose during the denoising process.

\begin{figure}[t]
\centering
\includegraphics[width=1\columnwidth]{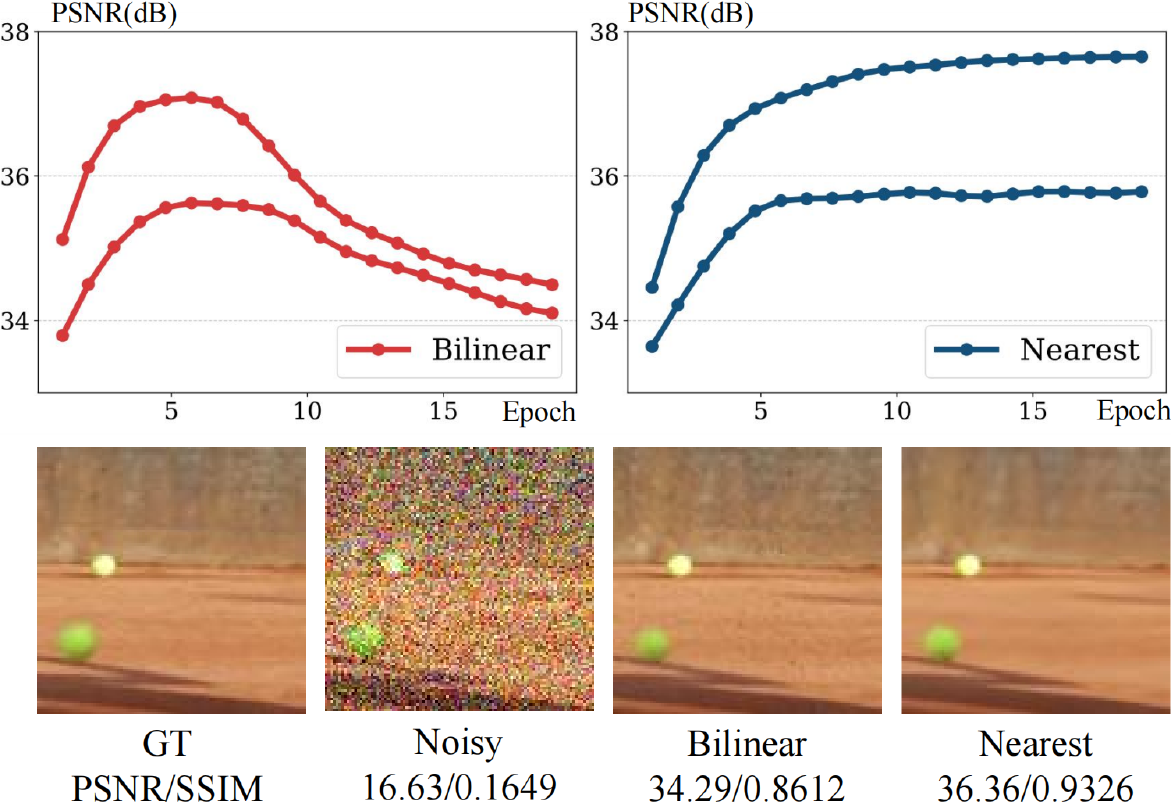} 
\caption{Visualizations of experimental results during training with different warping methods.}
\label{warp}
\end{figure}

\begin{figure}[t!]
\centering
\includegraphics[width=1\columnwidth]{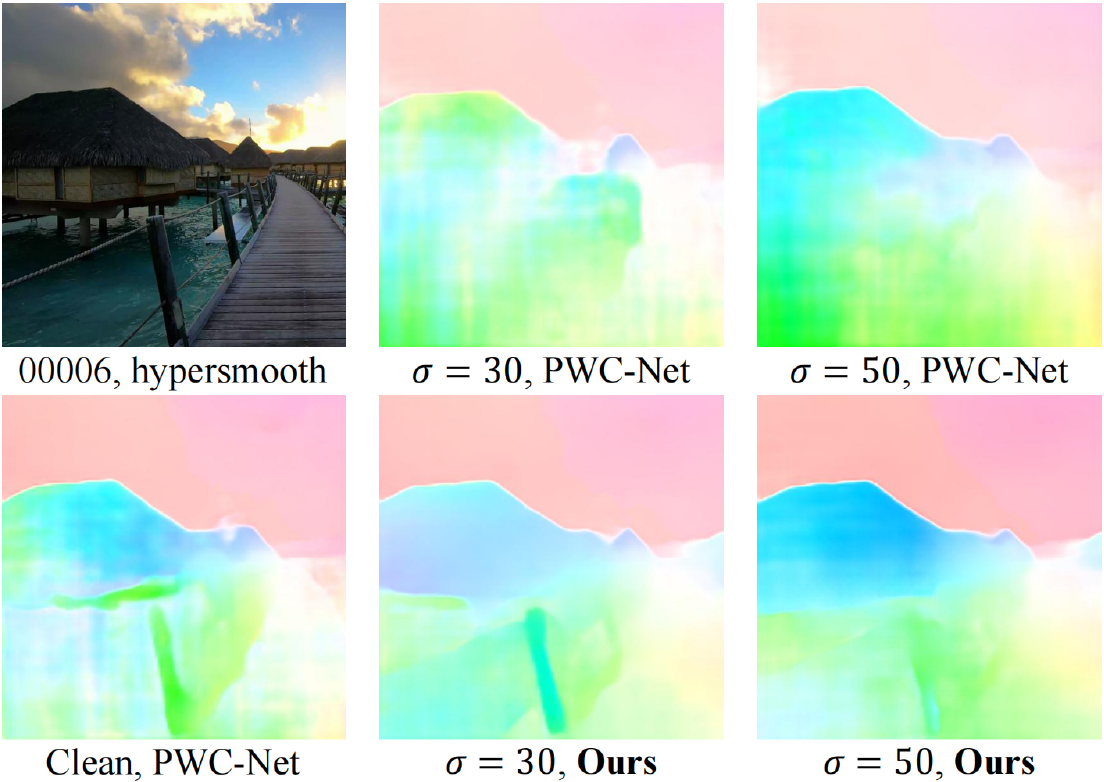}
\caption{Visualization of optical flow for the initial estimator compared to our refined results.}

\label{flow}

\end{figure}

\subsection{Analysis of the Proposed Method}
\subsubsection{Ablation study.}
We perform ablation studies on the Set8 dataset with Gaussian noise level $\sigma$=30, as detailed in Table~\ref{ablation}. Starting with temporal feature propagation alone, incorporating the BSA block enhances temporal feature utilization, improving PSNR by 0.35 dB. Adding the SRFE module further leverages spatial information, resulting in an additional 0.19 dB increase in PSNR. Finally, introducing optical flow refinement provides a further improvement of 0.08 dB in PSNR. These results demonstrate that gradually utilizing temporal and spatial features and refining alignment incrementally enhances model performance.
\subsubsection{Feature Alignment Strategy.}
Figure \ref{warp} presents the results of experiments conducted on two samples from the Set8 dataset using bilinear interpolation and nearest-neighbor interpolation. The former led to a decrease in PSNR during training, which aligns with our conclusions that bilinear interpolation disrupts the noise structure, thereby violating our blind-spot assumption. Consequently, bilinear interpolation produced poor visual results.
\subsubsection{Optical Flow Refinement.}
We visualize the refined optical flow produced by our proposed method for noise levels $\sigma=30$ and $\sigma=50$ as shown in Figure \ref{flow}. The optical flow estimator benefits from knowledge distillation guided by generated pseudo-ground truths in the training process, leading to more accurate optical flow predictions under noisy conditions. Consequently, our alignment module achieves improved matching accuracy, which in turn contributes to the superior performance of our denoising model.

\begin{table}[t]
\centering
\resizebox{0.98\linewidth}{!} {

\begin{tblr}{
  cells = {c},
  cell{1}{2} = {c=4}{},
  vline{2} = {-}{},
  hline{1,8} = {-}{0.08em},
  hline{2,6} = {-}{},
}
Component          & Methods  &       &        &       \\
Propagation    & \checkmark & \checkmark & \checkmark & \checkmark \\
BSA Block          &       & \checkmark & \checkmark & \checkmark \\
SRFE Module        &       &       & \checkmark & \checkmark \\
Optical Refinement &       &       &        & \checkmark \\
PSNR               & 32.14 & 32.49 & 32.68  & \textbf{32.76} \\
SSIM               & 0.8942 & 0.9037 & 0.9068 & \textbf{0.9072}
\end{tblr}}
\caption{Ablation study of model components.}
\label{ablation}

\end{table}


\section{Conclusion}
In this paper, we introduce STBN for self-supervised video denoising. We validate and calibrate the multi-frame alignment paradigm within a self-supervised framework to ensure the global consistency of the noise prior, thereby mitigating training bias. Our proposed spatiotemporal blind-spot feature aggregation preserves long-range temporal dependencies and enhances spatial receptive fields for comprehensive global perception. Additionally, our unsupervised optical flow refinement reduces sensitivity to noise, improving the precision of spatiotemporal feature utilization. Experimental results demonstrate that our method surpasses existing unsupervised approaches and shows strong comparability to supervised methods, demonstrating great potential.
\section*{Acknowledgments}
This work is supported by the Shenzhen Science and Technology Project under Grant (JCYJ20220818101001004).

\bibliography{aaai25}

\clearpage

\newpage
\appendix


\begin{center}
\textbf{\Large Supplementary Material}
\end{center}

\section{Proof of Blind-Spot Assumption in Videos}

\subsection{Model Definition}
We assume that at the $t$-th frame $\boldsymbol{y}_t$, the receptive field for the $i$-th pixel $\boldsymbol{y}_{(t,i)}$, which acts as the blind-spot in our model, is denoted as $\boldsymbol{y}_{t,RF(i)}$. We define our model as the function as follows:
\begin{multline}
f\big(\boldsymbol{y}_{t,RF(i)}, warp(\boldsymbol{y}_k, \boldsymbol{O}_{k}); \boldsymbol{\theta}\big) = \boldsymbol{y}_{(t,i)}, \\
k \in \{1,2,\ldots,T\} \setminus \{t\}.
\end{multline}
In this formulation:
\begin{itemize}
    \item $\boldsymbol{\theta}$ represents the vector of model parameters to be optimized.
    \item $\boldsymbol{O}_{k}$ denotes the estimated optical flow between frames.
    \item $warp$ denotes the alignment operation applied to the frames.
\end{itemize}

\subsection{Training Objective}
The model is trained by minimizing the following empirical risk:
\begin{equation}
\mathop{\arg\min}_{\boldsymbol{\theta}} \sum_{t,i} \boldsymbol{L}\left(f\left(\boldsymbol{y}_{t,RF(i)}, \text{warp}(\boldsymbol{y}_k, \boldsymbol{O}_{k}); \boldsymbol{\theta}\right), \boldsymbol{y}_{(t,i)}\right).
\end{equation}

We use the $\mathcal{L}_2$ loss as an example. The empirical risk can be expressed as:
\begin{equation}
\mathcal{L}(\boldsymbol{\theta}) = \sum_{t,i} \left\| f\left(\boldsymbol{y}_{t,RF(i)}, warp(\boldsymbol{y}_k, \boldsymbol{O}_{k});\boldsymbol{\theta}\right) - \boldsymbol{y}_{(t,i)} \right\|_2.
\end{equation}

In this formulation, $\mathcal{L}(\theta)$ measures the squared difference between the predicted values from the receptive field and the blind-spot values. 

\subsection{Proof of Equivalence to Supervised Training}
First, we introduce the assumption of blind spot networks in the image domain~\cite{krull2019noise2void,xie2020noise2same}:
\begin{equation}
\mathop{\arg\min}_{\boldsymbol{\theta}}\sum_iL\left(f(\boldsymbol{y}_{RF(i)};\boldsymbol{\theta}),\boldsymbol{y}_i\right)~, 
\end{equation}
which is equal to the supervied loss:
\begin{equation}
\mathop{\arg\min}_{\boldsymbol{\theta}}\sum_iL\left(f(\boldsymbol{y}_{RF(i)};\boldsymbol{\theta}),\boldsymbol{x}_i\right)+c, 
\end{equation}
where c is a constant.

According to the above assumption, the training process can be expressed using the $\mathcal{L}_2$ loss as:
\begin{equation}
\begin{aligned}
    &\sum_{i}\left\| f\left(\boldsymbol{y}_{(RF(i))}; \boldsymbol{\theta}\right) - \boldsymbol{y}_{i} \right\|_2
    \\= &\sum_{i}\left\| f\left(\boldsymbol{y}_{(RF(i))}; \boldsymbol{\theta}\right) - \boldsymbol{x}_{i} \right\|_2+c.
\end{aligned}
\end{equation}

Further, we generalize the equation to the video blind-spot assumption. According to the blind-spot network assumption, the receptive field $RF(i)$ of blind-spot is related to the underlying ground truth but independent of the noise values.  Note that the warp term $warp(\boldsymbol{y}_k, \boldsymbol{O}_{k})$ as aligned frames, satisfies both characteristics (aligned without compromising the noise independence). Therefore, we incorporate $warp(\boldsymbol{y}_k, \boldsymbol{O}_{k})$ as part of the $RF(i)$ as follows:
\begin{equation}
\begin{aligned}
&\mathop{\arg\min}_{\boldsymbol{\theta}} \sum_{t,i} \boldsymbol{L}\left(f\left(\boldsymbol{y}_{t,RF(i)}, \text{warp}(\boldsymbol{y}_k, \boldsymbol{O}_{k}); \boldsymbol{\theta}\right), \boldsymbol{y}_{(t,i)}\right) \\
=& \mathop{\arg\min}_{\boldsymbol{\theta}} \sum_{t,i} \boldsymbol{L}\left(f\left(\boldsymbol{y}_{t,RF(i)}; \boldsymbol{\theta}\right), \boldsymbol{y}_{(t,i)}\right)  \\
=& \mathop{\arg\min}_{\boldsymbol{\theta}} \sum_{t}\sum_{i} \boldsymbol{L}\left(f\left(\boldsymbol{y}_{t,RF(i)}; \boldsymbol{\theta}\right), \boldsymbol{y}_{(t,i)}\right).  \\
\end{aligned}
\end{equation}

Each term $t$ satisfies the equivalence condition of the blind-spot assumption. Therefore, we have:
\begin{equation}
\begin{aligned}
&\mathop{\arg\min}_{\boldsymbol{\theta}} \sum_{t}\sum_{i} \boldsymbol{L}\left(f\left(\boldsymbol{y}_{t,RF(i)}; \boldsymbol{\theta}\right), \boldsymbol{y}_{(t,i)}\right) \\
=& \mathop{\arg\min}_{\boldsymbol{\theta}} \sum_{t}\sum_{i} \boldsymbol{L}\left(f\left(\boldsymbol{y}_{t,RF(i)}; \boldsymbol{\theta}\right), \boldsymbol{x}_{(t,i)}\right) + c \\
=& \mathop{\arg\min}_{\boldsymbol{\theta}} \sum_{t,i} \boldsymbol{L}\left(f\left(\boldsymbol{y}_{t,RF(i)}, \text{warp}(\boldsymbol{y}_k, \boldsymbol{O}_{k}); \boldsymbol{\theta}\right), \boldsymbol{x}_{(t,i)}\right) + c.
\end{aligned}
\end{equation}

This shows that our training process is conceptually similar to the supervised training process, where the empirical risk is minimized. The above equivalence further emphasizes the need for a detailed discussion on the use of optical flow alignment strategies, as highlighted in the main text.

\begin{table*}[!]
  \centering
  \resizebox{1\linewidth}{!} {
  \renewcommand{\arraystretch}{1.1} 
  
  \begin{tabular}{c|c|ccccc|c}
    \toprule
    \multirow{2}{*}{Category} & \multirow{2}{*}{Method} & \multirow{2}{*}{$\sigma=10$} & \multirow{2}{*}{$\sigma=20$} & \multirow{2}{*}{$\sigma=30$} & \multirow{2}{*}{$\sigma=40$} & \multirow{2}{*}{$\sigma=50$} & \multirow{2}{*}{avg} \\
    & &  &  &  &  &  &  \\
    \midrule
    Traditional & VBM4D        & 36.05/- & 32.19/- & 30.00/- & 28.48/- & 27.33/- & 30.81/- \\
    \midrule
    \multirow{5}{*}{Supervised} 
    & DVDnet        & 36.08/0.9510 & 33.49/0.9182 & 31.68/0.8862 & 30.46/0.8564 & 29.53/0.8289 & 32.29/0.8881 \\
    & FastDVDnet    & 36.44/0.9540 & 33.43/0.9196 & 31.68/0.8889 & 30.46/0.8608 & 29.53/0.8351 & 32.31/0.8917 \\
    & FloRNN        & 37.57/0.9639 & 34.67/0.9379 & 32.97/0.9138 & 31.75/0.8911 & 30.80/0.8696 & 33.55/0.9153 \\
    & VRT           & \underline{37.88/0.9630} & \underline{35.02}/0.9373 & \underline{33.35}/0.9141 & 32.15/0.8928 & 31.22/0.8733 & \underline{33.92}/0.9161 \\
    & RVRT          & 37.53/0.9626 & 34.83/\underline{0.9383} & 33.30/\underline{0.9173} & \underline{32.21/0.8981} & 3\underline{1.33/0.8800} & 33.84/\underline{0.9192} \\
    \midrule
    \multirow{5}{*}{Unsupervised} 
    & UDVD          & 36.36/0.9510 & 33.53/0.9167 & 31.88/0.8865 & 30.72/0.8595 & 29.81/0.8349 & 32.46/0.8897 \\
    & RDRF          & 36.67/0.9547 & 34.00/0.9251 & 32.39/0.8978 & 31.23/0.8725 & 30.31/0.8490 & 32.92/0.8998 \\
    & ER2R$_s$      & 37.55/- & 34.34/- & 32.45/- & 31.09/- & 30.05/- & 33.10/- \\
    & Ours          & 37.24/0.9594 & 34.41/0.9322 & 32.76/0.9072 & 31.57/0.8837 & 30.62/0.8608 & 33.32/0.9087 \\
    & \textbf{Ours$_s$}       & \textbf{38.38/0.9670 }& \textbf{35.48/0.9432} & \textbf{33.77/0.9212} & \textbf{32.54/0.9005} & \textbf{31.56/0.8803} &\textbf{ 34.35/0.9224} \\
    \bottomrule
  \end{tabular}}
  
  \caption{Quantitative comparison of PSNR/SSIM on the Set8 dataset for Gaussian denoising. The best results for all the compared methods are in \textbf{bold}, while second is \underline{underlined}. $e$ represents self-supervised training on each single video on the Set8 dataset.}
  \label{tab:gauss_set8}
\end{table*}

\begin{figure*}[t]
\centering

\includegraphics[width=0.99\textwidth]{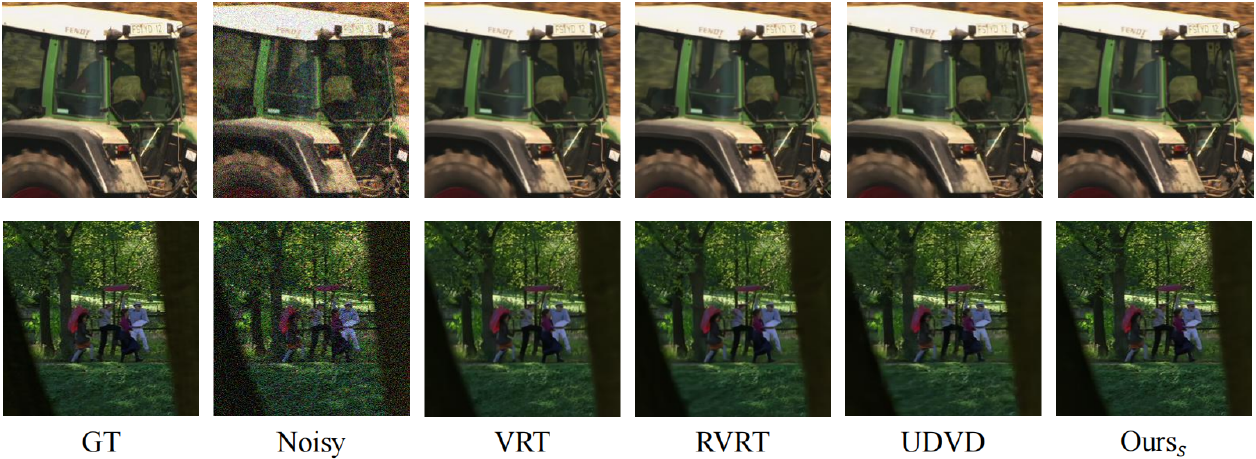} 
\caption{Visual comparisons on Set8 dataset. $s$ represents self-supervised training on each single video on the Set8 dataset.}
\label{vis:raw}
\end{figure*}

\section{Additional Model Analysis}

\subsection{Fully Self-Supervised on Test Set}
Self-supervised methods enable model optimization without the need for ground truth data. Several approaches~\cite{lee2022ap, zheng2023unsupervised} have demonstrated training and evaluation on test datasets to assess their models' performance under fully self-supervised conditions. Following these works~\cite{zheng2023unsupervised}, we conduct training and testing on the Set8 dataset.
\subsubsection{Implementation Details}
To generate noisy video sequences on Set8 dataset, additive white Gaussian noise (AWGN) with a standard deviation ${\sigma \in [5,55]}$ is introduced to the dataset.
We compare our method with a range of benchmarks, including the non-learning method VBM4D~\cite{maggioni2012video}, supervised approaches such as FastDVDnet~\cite{tassano2020fastdvdnet}, PaCNet~\cite{vaksman2021patch}, FloRNN~\cite{li2022unidirectional}, RVRT~\cite{liang2022recurrent}, VRT~\cite{liang2024vrt} as well as unsupervised methods like UDVD~\cite{sheth2021unsupervised}, RDRF~\cite{wang2023recurrent}, and ER2R~\cite{zheng2023unsupervised}. 

\subsubsection{Quantitative Comparison.} 
Table \ref{tab:gauss_set8} reports the PSNR and SSIM of different methods on the Set8 datasets under different noise levels. Under the conditions described above, our experiments achieve outstanding results, surpassing all existing SOTA methods in traditional, unsupervised and self-supervised categories. Specifically, under the same setting, our approach outperforms ER2R$_s$ by an average of 1.34 dB. This improvement is attributed to our bidirectional temporal propagation module, which leverages information from both forward and backward frames. Unlike ER2R, which only utilizes adjacent frames, lacks global perceptual capabilities. Additionally, we surpass the supervised SOTA method VRT by 0.43 dB. As a self-supervised method, our approach can be directly trained and applied to noisy video data, significantly enhancing practical utility and achieving notable performance gains.

\begin{table}[t]
\centering
\label{tab:res_set8}
\resizebox{0.9\linewidth}{!} {
\renewcommand{\arraystretch}{1.2} 
\begin{tabular}{c|c} 
\toprule
Methods & PSNR (dB)  \\ 
\hline
DIS~\cite{kroeger2016fast}     & 32.28  \\
SPyNet~\cite{ranjan2017optical}  & 32.41  \\
PWC-Net~\cite{sun2018pwc} & 32.68  \\
\bottomrule
\end{tabular}}
\caption{Ablation studies on Different Optical Flow Models.} 
\end{table}

\subsection{Hyperparameter of Modules}
We refine the optical flow estimation using the following loss function:
\begin{equation}
\mathcal{L}_{dis} = \sum_{t} \left\|\boldsymbol{\tilde{O}}_{t}^f - \boldsymbol{O}_{t}^f\right\|_1.
\end{equation}
This distillation loss, scaled by a small coefficient $\alpha$, is incorporated into the overall training process as a regularization term. Specifically, the refined optical flow $\boldsymbol{\tilde{O}}_{t}^f$ is obtained as follows:
\begin{equation}
\boldsymbol{\tilde{O}}_{t}^f=sg(\mathcal{E}_{\text{fix}}(\boldsymbol{\tilde{x}}_{t},\boldsymbol{\tilde{x}}_{t+1})),
~\boldsymbol{O}_{t}^f=\mathcal{E}(\boldsymbol{y}_{t},\boldsymbol{y}_{t+1}),
\end{equation}
where $sg(\cdot)$ denotes the stop gradient operation. The refined flow $\boldsymbol{\tilde{O}}_{t}^f$, treated as a pseudo-ground truth, is then used to guide the optical flow estimation process, enhancing accuracy in noisy video sequences. In practice, we follow by~\cite{sun2018pwc} to use the L2 norm to regularize parameters of the model:
\begin{equation}
\mathcal{L}_{dis} = \sum_{t} \left\|\boldsymbol{\tilde{O}}_{t}^f - \boldsymbol{O}_{t}^f\right\|_1+
\gamma|\Theta|_2.
\end{equation}
The distillation loss is introduced after the first 1,000 iterations of training with $\alpha=5 \times 10^{-4}$. 

As illustrated in Figure~\ref{flow}, we conduct ablation studies on the video \textit{hypersmooth} of Set8 dataset to determine the optimal selection of hyperparameters. Our ablation studies reveal a distinct trend in the PSNR metric as the hyperparameter value increases. Specifically, the PSNR metric exhibits a trend where it initially increases and then decreases, reaching its peak at a magnitude of $5 \times 10^{-4}$. This indicates that at this parameter setting, the optical flow refinement module aligns most effectively with our denoising model. The superior performance at this value suggests that the refined optical flow prediction, achieved under these conditions, enhances the spatial alignment, leading to a significant improvement in denoising performance. This result demonstrates the critical role of well-tuned hyperparameters in optimizing the synergy between optical flow refinement and denoising processes.
\subsection{Different Optical Flow Models}
Various optical flow models exhibit differing capabilities in capturing and predicting motion within sequences, a key factor that influences the effectiveness of video processing tasks such as denoising. These models differ significantly in their architectural designs, computational efficiency, and accuracy, each bringing distinct strengths and weaknesses to the table. By evaluating the characteristics of different optical flow models, we can better understand their potential impact on the denoising process and make informed decisions to enhance both the quality and efficiency of our approach.

To identify the most suitable optical flow model for our self-supervised video denoising method, we conduct a comparative analysis of three distinct optical flow estimation methods: the traditional approach, DIS~\cite{kroeger2016fast}, and learning-based methods, SPyNet~\cite{ranjan2017optical} and PWC-Net~\cite{sun2018pwc}. Each of these methods offers unique characteristics that influence their integration with our denoising framework. The traditional approach, while established, may lack the adaptability and precision of more modern techniques. DIS, known for its speed, provides a rapid yet reasonably accurate estimation of optical flow, making it a potential candidate for scenarios where computational resources are limited. On the other hand, learning-based methods like SPyNet and PWC-Net leverage deep learning to enhance flow estimation accuracy, albeit with varying degrees of computational demand. By assessing how well each of these methods integrates with our denoising framework, we aim to optimize the balance between denoising quality and computational feasibility, ultimately improving the overall performance of our approach.

\begin{figure}[t!]
\centering
\includegraphics[width=1\columnwidth]{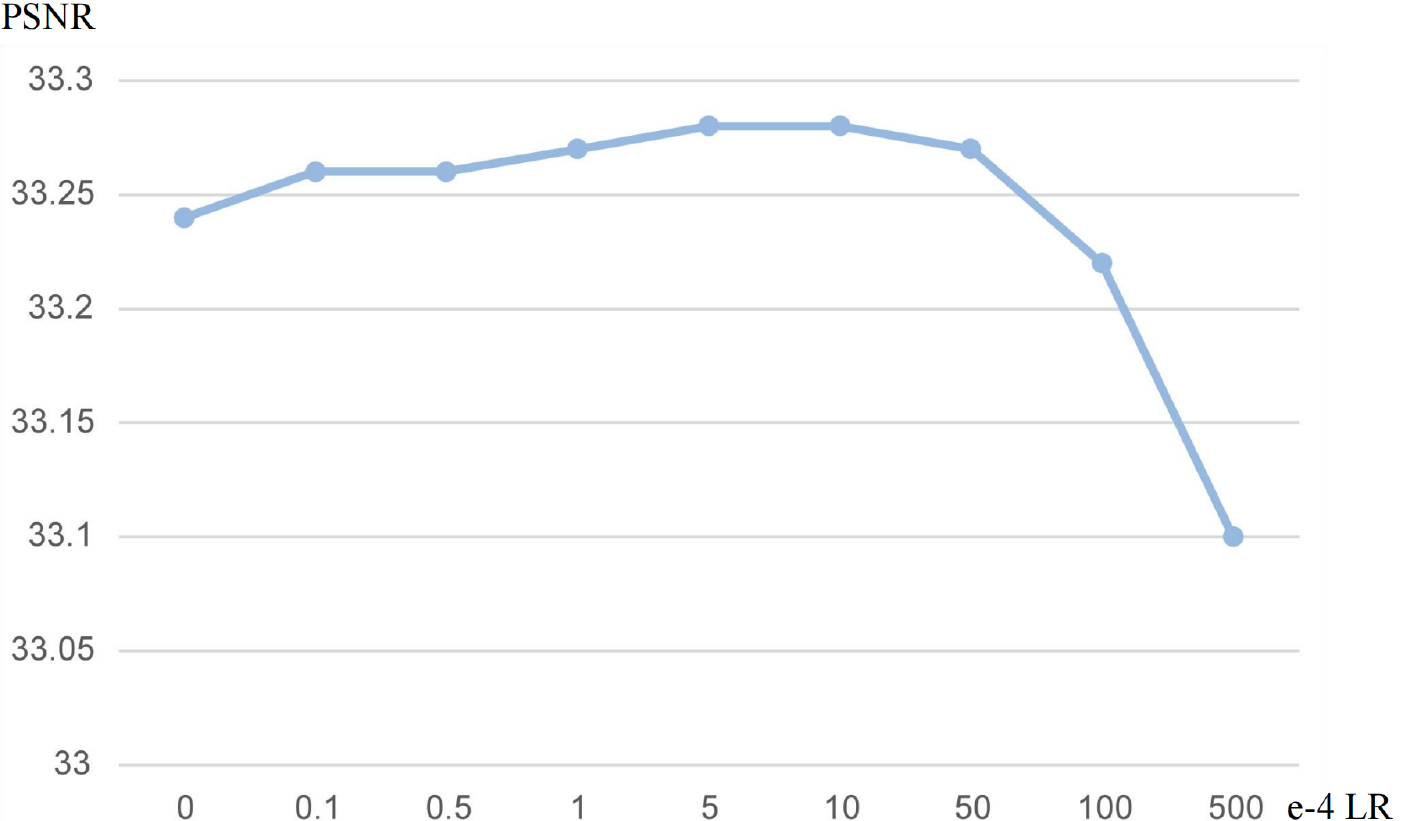}
\caption{Visualization of the impact of hyperparameters of optical flow refinement.}
\label{flow}
\end{figure}

\begin{figure*}[t]
\centering
\includegraphics[width=0.95\textwidth]{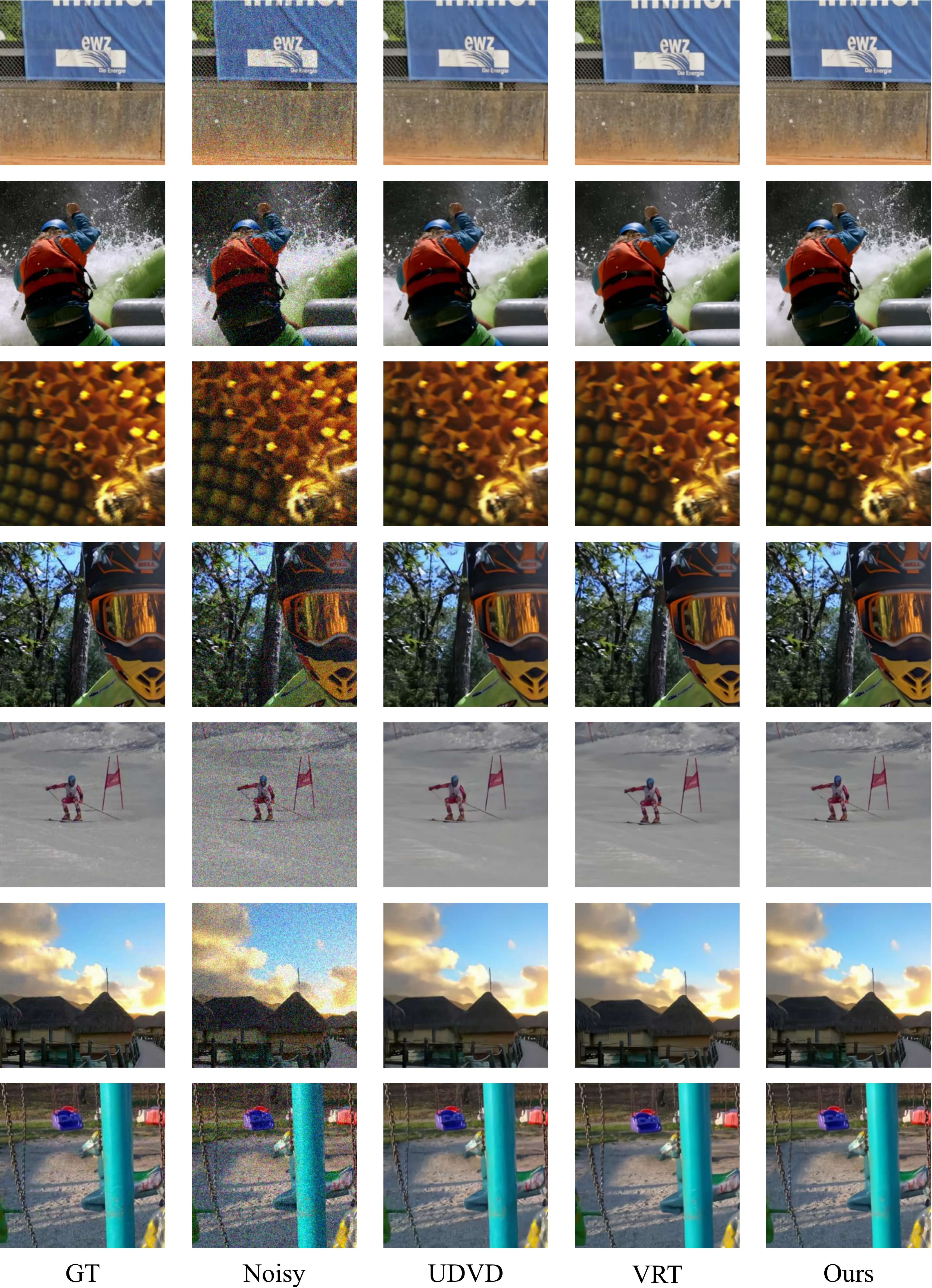} 
\caption{Visual comparisons of additional visual results.}
\label{vis}
\end{figure*}

\section{Additional Visual Results}
Additional visual results are provided to further illustrate the effectiveness of our approach in Figure~\ref{vis}. These results highlight the qualitative improvements achieved by our method, showcasing its ability to preserve finer details and reduce noise more effectively compared to existing techniques. By presenting these visual comparisons, we aim to offer a more comprehensive evaluation of our model’s performance across various challenging scenarios.

\end{document}